\documentclass[12pt,a4paper]{article}
\usepackage{csquotes}
\usepackage{booktabs}
\usepackage{adjustbox}
\usepackage{placeins}
\usepackage{float}
\usepackage{latexsym}
\usepackage{epsfig,longtable,lscape}
\usepackage[super,comma]{natbib}                                                   
\usepackage[table]{xcolor}
\usepackage{amsmath,amsfonts,euscript} 
\usepackage{amssymb}
\usepackage{verbatim,alltt}
\usepackage{float}
\usepackage{psfrag}
\usepackage{color}
\usepackage{times}
\usepackage{rotating}
\usepackage[normalem]{ulem}
\usepackage[T1]{fontenc}
\usepackage{mathptmx}
\usepackage{bigdelim}
\usepackage{booktabs}

\allowdisplaybreaks[4]
\usepackage{longtable,lscape}
\usepackage{hyperref}
\hypersetup{
	pdftitle = {},
	pdfauthor = {},
	pdfproducer = {},
	pdfcreator = {},
	pdfkeywords = {},
}

\begin{document}
	
	\long\def\comment#1{}  
	\def\p{\partial}
	\def\ds{\displaystyle}
	\def\sc{\scriptscriptstyle}
	\def\vdot{\mbox{\large \bf .}}
	
	\newcommand{\Frac}[2]{{\displaystyle #1 \over \displaystyle #2}}
	\newcommand{\bfmath}[1]{\mbox{\boldmath$#1$}}
	\newcommand{\fs}[2]{\fontsize{#1}{#2}\selectfont}
	\newcommand{\revrea}[2]{ \overset{\ds #1}{\underset{\ds #2}{\mbox{\scalebox{2}{$\rightleftharpoons$}}}} }

	\title{%
		\textbf{
			A Novel Method Combines Moving Fronts, Data Decomposition and
			Deep Learning to Forecast Intricate Time Series} \\
		\large [PRE-PRINT]
	}

\author{Debdarsan Niyogi \footnote{\scriptsize{E-mail:
						debdarsan.niyogi@gmail.com,
						ORCID: 0000-0002-2376-5482}} \\
			Divecha Centre for Climate Change, \\
			Indian Institute of Science,
			Bangalore 560012, India }

		\maketitle
	

	
	\begin{abstract}
		A univariate time series with high variability can pose a challenge even to Deep Neural Network (DNN). To overcome this, a univariate time series is decomposed into simpler constituent series, whose sum equals the original series. As demonstrated in this article, the conventional one-time decomposition technique suffers from a leak of information from the future, referred to as a data leak. In this work, a novel Moving Front (MF) method is proposed to prevent data leakage, so that the decomposed series can be treated like other time series. Indian Summer Monsoon Rainfall (ISMR) is a very complex time series, which poses a challenge to DNN and is therefore selected as an example. From the many signal processing tools available, Empirical Wavelet Transform (EWT) was chosen for decomposing the ISMR into simpler constituent series, as it was found to be more effective than the other popular algorithm, Complete Ensemble Empirical Mode Decomposition with Adaptive Noise (CEEMDAN). The proposed MF method was used to generate the constituent leakage-free time series. Predictions and forecasts were made by state-of-the-art Long and Short-Term Memory (LSTM) network architecture, especially suitable for making predictions of sequential patterns. The constituent MF series has been divided into training, testing, and forecasting. It has been found that the model (EWT-MF-LSTM) developed here made exceptionally good train and test predictions, as well as Walk-Forward Validation (WFV), forecasts with Performance Parameter ($PP$) values of 0.99, 0.86, and 0.95, respectively, where $PP$ = 1.0 signifies perfect reproduction of the data.
	\end{abstract}
	
	\subsubsection*{Keywords} Time series forecast, EWT, decomposition, LSTM, Indian Summer Monsoon Rainfall
	
	\section*{Introduction}Machine Learning (ML) techniques have achieved significant successes
	in the prediction of univariate time series such as inter-annual variation of rainfall during Indian
	Summer Monsoon (ISMR). ML models are developed by dividing the time series
	into a training set and a testing set. The model developed by using
	the former is evaluated against the latter and evolves by
	a repetition of these steps until the desired accuracy for testing data is reached.  It is important to keep 
	the test data set separate from the model building during
	training. If this is not carefully enforced,  testing of the
	performance of the model will not be rigorous.  Many ML models are often based on the reduction of the variability of the time series by distributing
	it among simpler constituents, using signal decomposition
	methods. The decomposition is often performed on the entire data
	set once and then the training and testing sets are generated by dividing the constituent series. This methodology 
	suffers from data leakage, as will be discussed later, which makes ML model building ineffective. Hence, models so built will be unfit for making forecasts. A method
	is developed in this paper to prevent data leakage, enabling the usage of the subseries in conventional methods. 
	
	The proposed method decomposes the complex time
	series, by using Empirical Wavelet Transform (EWT), into a finite number of constituent functions to make machine learning
	easier. However, the innovation is in the novel method
	which collates information from Moving Fronts (MF), to get rid of the data leak. The data from 
	constituent functions obtained from MF were used as parallel series, to form a multivariate problem. This
	was synthesized with deep learning techniques using
	LSTM network. 
	
	Prediction of inter-annual variation of ISMR is important
	and this was taken up for a demonstration of the method proposed.
	Many attempts have been made
	over a long period to predict the ISMR.
	A large number of articles have also been written based on statistical
	methods, regression techniques, ANN, DNN modelling and mixtures of different
	methods also have been proposed. 
	All these have achieved only limited success in both testing and forecasting, and this series was chosen here to demonstrate
	the efficacy of the method proposed.
	Free from any future information leakage, the results obtained both for testing and forecasting
	in the present work are very good.
	
	The article has been organized as follows. Decomposition techniques and data leak is discussed first. A survey of 
	research work ISMR that employ decomposition is presented next. It is followed by an exploratory
	data analysis, and a discussion of the method
	to avoid data leaks.  The results obtained are discussed and conclusions follow at the end.
	
	\section*{Previous Related Work}
	\subsection*{Decomposition of Time Series and Data Leak}
	Many published
	works on time series from a variety of disciplines employ
	Empirical Mode Decomposition (EMD) or its variants, Empirical
	Wavelet Transform (EWT),  Variational Mode Decomposition (VMD) to decompose the entire available data set and then
	divide the resulting components into training and testing
	~\cite{Zhang2008,Bai2009,Liu2012,Wei2012,Ghelardoni2013,Ren2015,Wang2016,Zhang2016,
		Huang2021, Bisoi2019, Li2019, E2017, Zhou2019, Yang2019,
		Lahmiri2017}. This is referred to as one-time
	decomposition. However, two deficiencies have been detected
	in this strategy of one-time decomposition. Firstly,  there is a scope
	for information from the test period percolating into the training period,
	and hence testing of the model is not rigorous. Secondly, it does not work well in
	forecasting when the range of time series is extended beyond the last data
	point of the testing set. This in part can be attributed
	to the weakness in testing resulting from the transfer of information
	just now referred to. 
	
	\citet{Wu.wang2016}, using EMD drew attention to the
	fact that every addition of a new data point to an existing
	time series significantly alters the sub-series after
	decomposition including the new point. Therefore, adding 
	future data changes the original training and testing
	data. They surmised that this could be the flaw that prevents
	successful forecasting.  
	
	\citet{Gao_2021} applied 
	EWT to open datasets, and clearly shows the problem of data
	leakage graphically using data. They pointed out that
	the practice of one-time decomposition causes data leakage
	from future values.  They used Random Vector Functional Link
	(RFVL) as a machine learning model, and EWT. A walk-forward
	technique had to be used to obtain good results.
	
	\citet{Huang2021} also pointed out the drawback of data leaks. They used VMD with
	moving windows to eliminate data leaks.
	
	The problem of possible data leaks was recognized by workers attempting
	to predict and forecast ISMR. \citet{Iyengar2004} proposed
	the decomposition of ISMR into EMD. However, the range of data was such
	that only one mode function needed a neural network to predict while all
	the others could be fitted by regression. They could forecast but only
	for a short period of about ten years. \citet{Johny2020} decomposed the ISMR data by Ensemble Empirical Mode
	Decomposition (EEMD). They recognized the data leak issue and could
	forecast by a walk-forward technique that required the development of new
	ANN model whenever a new data point was added.
	
	It is thus clear that data leak is a problem that needs to
	be circumvented to develop better artificial intelligence
	models to make predictions and forecasts of complex time series.

	\subsection*{Prediction and Forecasting of ISMR}
	A performance parameter ($\mathit{PP}$) is
	defined here, and is used  to give the level
	of predictive skill reached in the work reviewed.
	It is defined as
	\[PP = 1 - {\left( RMSE / SD \right)}^2\]
	where RMSE is the Root Mean Squared Error and SD is the Standard
	Deviation of the data.
	$\mathit{PP}$  $>$ 0 indicates a better prediction than the mean
	value. The closer the $\mathit{PP}$ is to 1, the
	better the forecast. As $PP$ gives a better insight into how much the RMSE of the prediction
	differs from the standard deviation (SD) of the observations, this has
	been selected as the measure of the accuracy of the prediction in this
	work as well.
	The voluminous work done on ISMR is skipped and only the works that involved decomposition are discussed.		 
	Work unrelated to ISMR is mentioned briefly
	for completeness.
	\subsection*{Hybrid Methods}
	A hybrid method consisting of ANN/LSTM in conjunction with data decomposition is used in this
	work. Previous related works are reviewed to indicate the usefulness
	of the hybrid methods. Work unrelated to ISMR is mentioned briefly
	for completeness.
	
	\subsubsection*{ANN Accompanied by Wavelet Decomposition}Here, data
	are decomposed using Discrete  
	Wavelet Transform (DWT) instead of EMD. This method has been widely used in forecasting run-off and flood
	forecast, {\it e.g.,} see paper by \citet{Sahay2013} and literature
	cited there. 
	
	\citet{VenkataRamana2013} applied this combination of Wavelet and ANN
	models (WANN) to the monthly rainfall data of the Darjeeling rain gauge
	station. They concluded that the hybrid method worked better than the ANN
	models.  \citet{Azad2015} focused on the most significant Spectrally
	Homogeneous Region (SHR), which has characteristics of South West
	monsoon but to analyse variance. They concluded that a hybrid model
	could account for a 45\% variance in the observed rainfall data. 
	
	\subsubsection*{Empirical Mode Decomposition (EMD) and ANN}
	\citet{Iyengar2004} proposed that the ISMR time series be decomposed
	into uncorrelated empirical modes called Intrinsic Mode Functions
	(IMF). Method of \citet{Huang1998} was used for decomposition. ISMR
	data between the years 1871-2002 was considered. The first IMF
	accounted for the highest variability and was modelled using ANN. The
	remaining IMFs were amenable to linear auto-regression. The sum of all
	the predicted IMF values constituted the forecast. The forecast was made
	for the years 1991-2002 with a value of $\mathit{PP}_{Test}$ = 0.82.
	The introduction of this idea of decomposing the time series into less
	variable constituent time series had been adapted by
	others. \citet{BeltrnCastro2013} followed this method to analyse
	rainfall data at Manizales city, Colombia. \citet{Basak2020} also used
	this methodology to analyse South West Monsoon rainfall data of the
	Gangetic West Bengal (GWB), but used Generalized Regression Neural
	Network (GRNN).  The $\mathit{PP}_{Test}$ value reported as 0.75. 
	
	\citet{Johny2020} decomposed the ISMR data by Ensemble Empirical Mode
	Decomposition (EEMD). They pointed out that decomposing the entire
	data using EEMD and then dividing the decomposed data into training
	and testing will use information from the future, and hence the forecast
	will be unreal. They introduced a new forecasting technique called
	Adaptive Ensemble Empirical Mode Decomposition-Artificial Neural
	Network (AEEMD-ANN) to overcome this problem. The data were
	partitioned into training (1871-1970) and testing (1971-2014). The
	training data were decomposed into IMFs and a residue.  ANN models
	were developed from training, and a one-step forecast was made using
	those models. Then data from the testing set, are added to the
	training set one at a time. After each addition, the training data is
	re-decomposed, and new models were developed to make the next
	forecast. They presented parity plots, but $PP$ for the forecasts could be calculated from the data presented and was found to be 0.75.

	The present work strives to achieve superior predictions
	using an approach that hybridizes controlled signal decomposition by EWT with
	a state-of-the-art sequence learning network (LSTM) that makes predictions based on its memory of historical data.
	
	\section*{Methodology of Processing Data}
	\subsection*{Exploratory Data Analysis}
	The rainfall data from the year 1871 to 2016 and 2017 to 2022 
	were obtained from the
	Indian Institute of Tropical Meteorology (IITM) and Indian Meteorological Department's (IMD) websites respectively.\footnote{Data were provided by Prof J Srinivasan, Divecha Center for Climate Change, Indian Institute of Science, Bangalore}. The summer rainfall,
	a sum of rainfall in June, July, August and September (will be
	termed as JJAS henceforth), has been chosen to make predictions. 
	
	The descriptive statistics of the rainfall data are presented in
	below Table \ref{table:descstat}. 
	
	\begin{table}[h!]
		\centering
		\begin{tabular}{| l | l |} 
			\hline
			Measure & Value \\
			\hline
			Data Points & 152 \\
			Average & 850.01 \\
			Std. deviation (SD) & 83.16 \\
			Max & 1020.2 \\
			Min & 604.0 \\
			\hline
		\end{tabular}
		\caption{Descriptive statistics of the rainfall data, 1871 to 2022}
		\label{table:descstat}
	\end{table}
	The probability density function revealed a skewed
	distribution. Almost no (or very weak) autocorrelation was
	found, and this explains why the linear time-series modelling
	failed. Seasonality was found to be absent. The trend was also found to
	be absent using the Augmented Dickey-Fuller (ADF) test of
	\citet{SAID1984}.  Hence, the data, as retrieved, were used as the input
	to the predicting models. Details are presented in Supplementary
	information A.
	
	\subsection*{The Choice of the Decomposition Technique}
\begin{figure}[H]
		\begin{center}
			\includegraphics[width=0.55\linewidth, keepaspectratio]{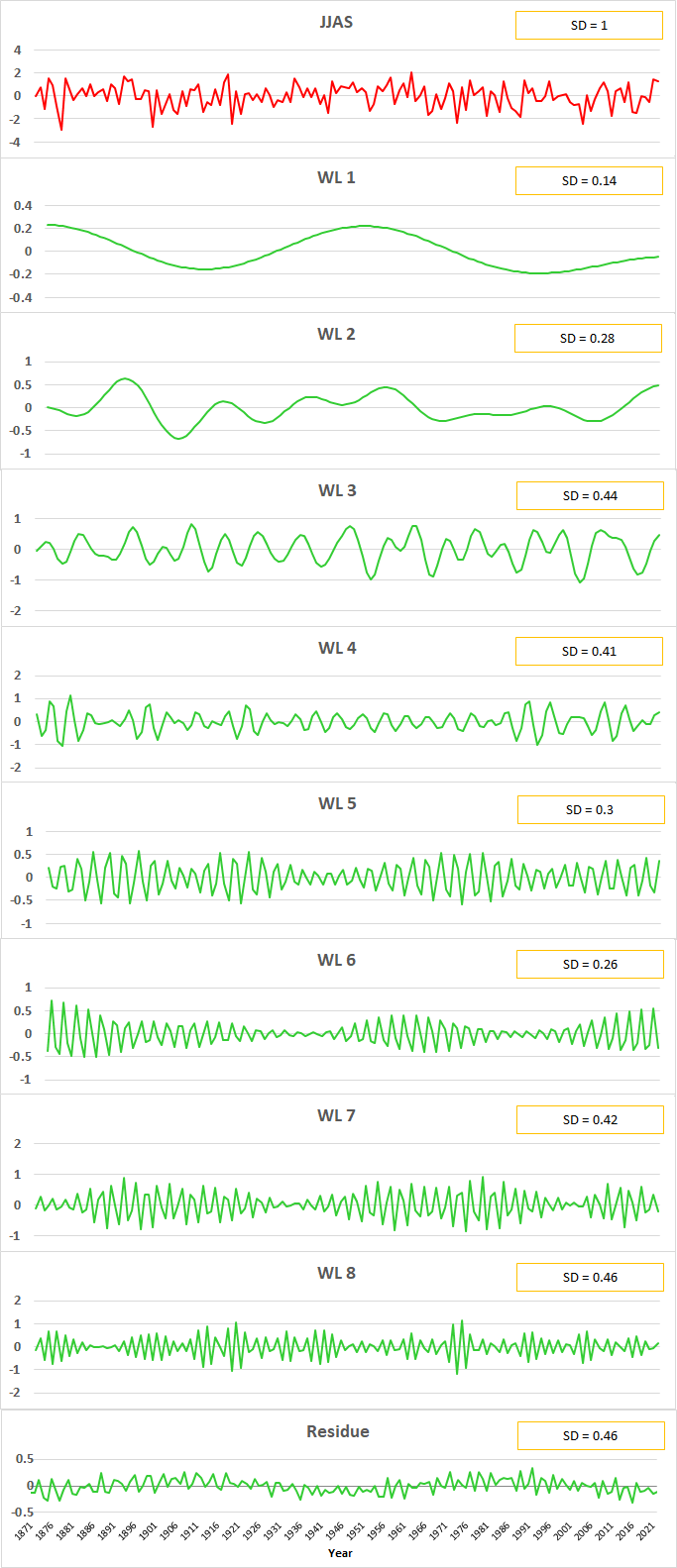}
			\caption{The component time series (labelled WL) obtained by applying EWT
				\label{fig:ewtwls}}
		\end{center}
	\end{figure}

	As mentioned earlier, one way to improve prediction accuracy is to
	distribute the variability of the time-series data into constituent
	child sub-series.  To achieve this, many signal decomposition methods
	have been proposed.  There are several methods based on Empirical Mode
	Decomposition proposed by \citet{Huang1998}. For all those, the user has no control
	over the number of modes that result after decomposition. Several
	improvements were made to the original method, and the advanced
	one is a Complete Ensemble Empirical Mode Decomposition with Adaptive
	Noise (CEEMDAN).  Here,  CEEMDAN was used as a representative of EMD
	techniques. The other widely used method is EWT.  The user can
	define the number of modes to be extracted in EWT,
	whereas there is no control over the number of IMFs generated by the EMD
	family. This can offer an advantage since by specifying a
	higher number of modes, one can obtain a more uniform distribution of
	the variability in the time series. The investigation into the distribution of variability of the modes indicated that
	the complexity is accumulated in IMF1 obtained from CEEMDAN.  Details of this work are presented in Supplementary
	information C.  EWT with the higher number of modes was
	tried as an alternative to decompose JJAS. It was found to be
	suitable for this work. The component time series obtained by EWT decomposition is shown in Figure \ref{fig:ewtwls}.
	
	\section*{New Scheme to Eliminate Data Leak}Before we proceed to discuss the new scheme developed in this work, it is necessary to draw attention to
	an inherent difficulty in the use of decomposed data for predicting/forecasting. 
	\section*{Decomposed Data and Information Leak}
	As mentioned earlier, only if the test data
	set is kept carefully aside and unseen by the model, the comparison of predictions with observed values will be an
	effective test of the performance of the model. The pre-requisite for this
	scheme to work is that the training and the testing data must remain
	constant so that patterns detected by DNN are fixed. Further,
	there should be no percolation of information from the test
	set to the training set.   The data leakage problem during
	one-time decomposition has already been mentioned.  Our investigations with JJAS decomposed using the whole
	data set also confirmed these findings. An analysis of this problem is presented next and is followed
	by a proposed solution.
	
	\subsection*{Use of Information from Future}
	The following observation indicates that  leak or sharing of information from
	testing to training set can occur.  
	When a data set is decomposed into component series, the 
	resultant components can change appreciably even when one new
	data point is added, {\it e.g.,} when the period of JJAS is changed
	from 1871-2000 to  1871-2001.  This is described below.
	Let $\mathbb{J}$ be the variable to be predicted from the JJAS data series:
	
	$$\mathbb{J} = \{J_{y_1}, J_{y_2}, J_{y_3}, ..., J_{y_N}\}$$
	where $J_{y_i}$ is the rainfall data for the year $y_i$. In the
	present case, $y_1$ = 1871 and $y_N$ = 2022 for the entire data
	set. We can model the entire set or subsets of it. We 
	will consider a general case where only a part of the series may be
	used in calculations. Let the series $\mathbb{J}$ from $y_n$ to $y_m$
	be decomposed into $k$ component series where  $y_n$ and $y_m$ are the first and end
	years. The data for each component is a value corresponding to a year. Thus, data for the year $y_l$ will
	be the values denoted by $\mathbb{C}_{y_n, y_m}(y_{l}, 1),   \mathbb{C}_{y_n,
		y_m}(y_{l}, 2), \ldots \mathbb{C}_{y_n, y_m}(y_{l}, k)$. Subscripts
	indicate a range of data while arguments indicate the year and number of
	the constituent series.  The entire
	decomposed data of the component series is then a matrix having dimension
	$(y_m - y_n) \times k$.
	Let $\mathbb{C}_{y_n, y_m}$ denote the matrix, and the elements are given by $\mathbb{C}_{y_n, y_m}(y_j,l)$,
	where\, $j = n\, \text{to}\, m$, and\, $l=1\, \text{to}\, k$.
	
	\[
	\mathbb{C}_{y_n, y_m} = 
	\begin{Bmatrix}
		$$\mathbb{C}_{y_n, y_m}(y_n, 1) & \mathbb{C}_{y_n, y_m}(y_n, 2) & ... & \mathbb{C}_{y_n, y_m}(y_n, k)$$ \\
		$$\mathbb{C}_{y_n, y_m}(y_{n+1}, 1) & \mathbb{C}_{y_n, y_m}(y_{n+1}, 2) & ... & \mathbb{C}_{y_n, y_m}(y_{n+1}, k)$$ \\
		\vdots \\
		$$\mathbb{C}_{y_n, y_m}(y_{m-1}, 1) & \mathbb{C}_{y_n, y_m}(y_{m-1}, 2) & ... & \mathbb{C}_{y_n, y_m}(y_{m-1}, k)$$ \\
		$$\mathbb{C}_{y_n, y_m}(y_{m}, 1) & \mathbb{C}_{y_n, y_m}(y_{m}, 2) & ... & \mathbb{C}_{y_n, y_m}(y_{m}, k)$$
	\end{Bmatrix}
	\]
	Each column of $\mathbb{C}_{y_n,y_m}$ forms a data series, and is like
	a conventional time series.
	
\begin{figure}[H]
		\begin{center}
			\includegraphics[width=.9\textwidth]{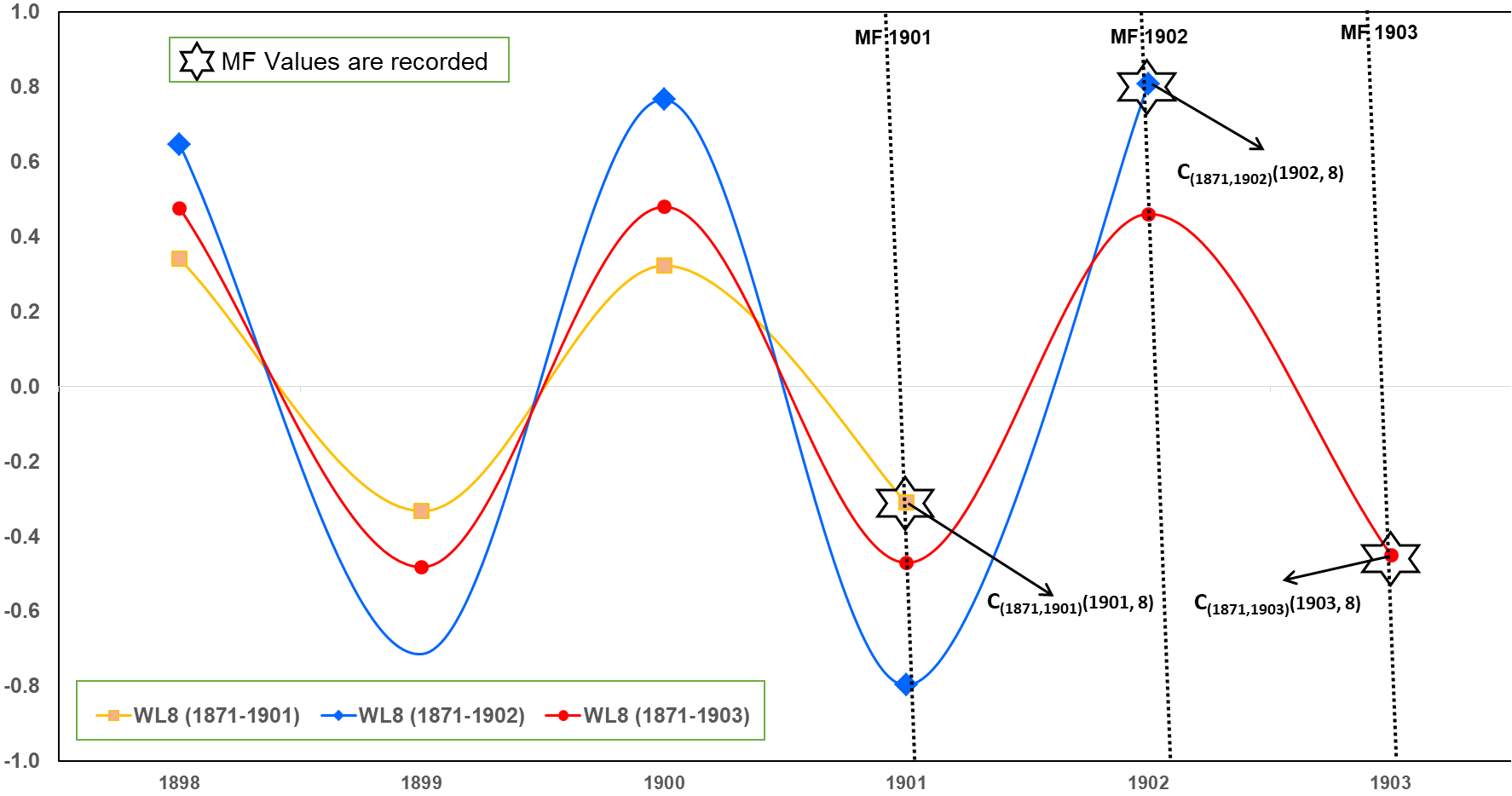}
			\caption{Variation of component labelled WL8 with Progressive Decomposition, Starting with 1901 and ending at 1903
				\label{fig:WF-1901-1903}}
		\end{center}
	\end{figure}

	When the period of the decomposition changes,  {\it even when the starting year $y_n$
		is fixed,} all the individual elements of the data series, {\it i.e.,} of each of the
	columns, can change. Thus, if the decomposition is done from year $y_n$
	to $y_{m+1}$,  an additional row
	comprising of $\mathbb{C}_{y_n, y_{m+1}}(y_{m+1}, l)$ is added to the
	matrix. The matrix will now be:
	
	\[
	\mathbb{C}_{y_n, y_{m+1}} = 
	\begin{Bmatrix}
		$$\mathbb{C} _{y_n, y_{m+1}}(y_n, 1)& \mathbb{C} _{y_n, y_{m+1}}(y_n, 2) & ... & \mathbb{C}_{y_n, y_{m+1}}(y_n, k)$$ \\
		$$\mathbb{C} _{y_n, y_{m+1}}(y_{n+1}, 1) & \mathbb{C}_{y_n, y_{m+1}}(y_{n+1}, 2) & ... & \mathbb{C}_{y_n, y_{m+1}}(y_{n+1}, k)$$ \\
		\vdots \\
		$$\mathbb{C}_{y_n, y_{m+1}}(y_{m-1}, 1) & \mathbb{C} _{y_n, y_{m+1}}(y_{m-1}, 2) & ... & \mathbb{C} _{y_n, y_{m+1}}(y_{m-1}, k)$$ \\
		$$\mathbb{C} _{y_n, y_{m+1}}(y_{m}, 1) & \mathbb{C} _{y_n, y_{m+1}}(y_{m}, 2) & ... & \mathbb{C}_{y_n, y_{m+1}}(y_{m}, k)$$ \\
		$$\mathbb{C}_{y_n, y_{m+1}}(y_{m+1}, 1) & \mathbb{C}_{y_n, y_{m+1}}(y_{m+1}, 2) & ... & \mathbb{C}_{y_n, y_{m+1}}(y_{m+1}, k)$$
	\end{Bmatrix}
	\]
	\noindent where the elements of this matrix for the  first $m$ rows can have
	values different from those in the matrix obtained from decomposition
	of the previous year's data:  $\mathbb{C} _{y_n, y_m}(y_j, l) \neq
	\mathbb{C}_{y_n, y_{m+1}}(y_j, l) $. This was found to be the
	case with 
	JJAS, and also by others.\cite{Gao_2021,Huang_2021}
	This behaviour is illustrated in Figure \ref{fig:WF-1901-1903} for one
	component labelled WL8. Three decompositions from JJAS data were made: 1871-1901, 1871-1902 and 1871-1903 (data
	shown from 1898). It shows clearly how the component labelled WL8 changes
	as data of one extra year are added. 
	
	{\it The implication of this is that information on the entire range of data
		is embedded in all the elements of the matrix obtained by the decomposition of data
		corresponding to all the years.} Thus, even if data for one additional
	year is added, the information of that year is incorporated into all
	the coefficients corresponding to earlier years, and the character of all the component series changes.  To be precise, the data
	$\mathbb{C}_{y_1, y_m}$ represent the state of the  component series
	as  would
	finally, be evolved to in the year $y_m$. This has two
	consequences. Firstly, suppose the data from $y_1$ to $y_m$ of the
	time-series ({\it e.g.,} 1871-1901) were divided
	into training set from $y_1$ to $y_n$  ({\it e.g.,} 1871-1891)  
	and the rest  $y_{n+1}$ to $y_m$ ({\it e.g.,} 1892-1901) as
	a testing set, and a model was developed. It could do very well to
	predict the test data, but that could as well be since information of
	$y_{n+1}$ to $y_m$ ({\it e.g.,} 1892-1901) was present in all the
	coefficients from $y_1$ up to $y_n$  as explained just now.  Thus, in one-time
	decomposition method, \textit{both} training and
	testing data are using information from 'future'. \citet{Johny2020}, 
	\citet{Gao_2021}, \citet{Huang_2021} had recognized this. It was termed \cite{Gao_2021}
	as  {\it {data-leakage}} problem. Secondly,
	the component series from $y_1$ to $y_n$ used for training but obtained from years $y_1$ to $y_m$ are
	very different from those for the same period ($y_1$ to $y_n$) but obtained
	from  $y_1$ to $y_{m+1}$ of the time series data. Thus, the prediction of JJAS for the
	year $y_{m+1}$ based on training on the earlier data
	set cannot be expected to succeed.  
	Thus, a forecast of data of JJAS even for one
	year beyond the last data point, {\it i.e.,} for  $y_{N+1}$,  cannot be expected to be successful
	based on training and testing of JJAS data from
	$y_1$ to $y_N$. While the procedure may succeed in some
	instances, it cannot be a reliable method.
	
	\subsection*{Proposed Strategy}
	Changes in the components of the component series ({\it i.e.,} elements of columns of the matrix
	$\mathbb{C}_{y_n, y_m}$) when a time series is decomposed
	after a new data point is  added to it are indicative of the
	difficulties pointed out.  These would be resolved if components of a series obtained by decomposition can be made to remain unchanged
	during  progressive decomposition, and only a  new element is
	added 
	to the existing data series with every decomposition.  If that can be achieved, a time series
	can be decomposed, and divided into training and testing sets while
	ensuring that information from the testing period is not permeating into
	training period. Then, the time-series
	can be modelled using
	the usual method of training and testing,
	accompanied with 
	the advantages of decomposition into simpler parallel data
	series that are more amenable to the learning process of the ML
	algorithms. A successful model can then be used for forecasting with
	some hope.
	
	We propose the following method to achieve the constancy desired. Some finite period is
	required to begin the decomposition of rainfall data into component series. Let that be
	$y_b$. Let JJAS data from years $y_1$ to  $y_b$  be decomposed into $k$
	component series. The rainfall for year $y_b$ is given by
	$$J_{y_b}= \sum_{l=1}^{k} \mathbb{C}_{y_1,y_b}(y_b,l) $$
	It is to be noted that $J_{y_b}$ is part of
	historical data and is constant, and hence
	$\mathbb{C}_{y_1,y_b}(y_b,l)$ are also constant.
	Now let data for one more year be added, and JJAS data from years $y_1$
	to  $y_{b+1}$ be decomposed into $k$ wavelets.   Before the rainfall
	for year $y_{b+1}$ is given by 
	$$J_{y_{b+1}}= \sum_{l=1}^{k} \mathbb{C}_{y_1,y_{b+1}}(y_{b+1},l) $$
	and the rainfall for $y_b$ is {\it also} given by
	$$J_{y_b}= \sum_{l=1}^{k} \mathbb{C}_{y_1,y_{b+1}}(y_b,l) $$
	As discussed earlier, $\mathbb{C}_{y_1,y_{b+1}}(y_b,l) \neq
	\mathbb{C}_{y_1,y_b}(y_b,l)$. 
	The key point of the proposed method is to recognize that {\it we do not
		have to use} $\mathbb{C}_{y_1,y_{b+1}}(y_b,l)$ to predict rainfall $J_{y_b}$, 
	but instead should use  $\mathbb{C}_{y_1,y_b}(y_b,l)$. The main
	advantage is that, as pointed out earlier,  $\mathbb{C}_{y_1,y_b}(y_b,l)$  are not affected by the 
	addition of data for $y_{b+1}$, and
	remain unchanged. Thus, we need to
	record only $\mathbb{C}_{y_1,y_b}(y_b,l)$ for the year $y_b$ and
	$\mathbb{C}_{y_1,y_{b+1}}(y_{b+1},l)$ for the year $y_{b+1}$. The argument can be extended for the years to
	follow. Thus, we propose that 
	coefficients of component functions for a year corresponding to the rainfall for
	that year only will be recorded. Those form a time series which can be learned by ML methods. 
	It can be imagined that each component of the series has a front, with
	the ending discrete value of the component of the series lying on this
	front. The front moves as more and more data of JJAS are added
	and decomposed. Let us denote the moving front by MF. Using  JJAS data, decomposed using EWT, the MFs are
	shown in Figure \ref{fig:WF-1901-1903} by vertical dotted lines. MF moves
	year by year, and the ending values at each MF are recorded (denoted by
	an asterisk in Figure \ref{fig:WF-1901-1903}). Between successive MFs,
	the data changes, but for any MF in a particular year, the stored
	end-point remains constant. Let $\mathbb{E}_{y_b,y_N}$ be the matrix
	of endpoints so recorded by decomposing JJAS data but beginning from the year $y_b$. It will be given by
	\[
	\mathbb{E}_{y_b, y_N} = 
	\begin{Bmatrix}
		$$\mathbb{C}_{y_1,y_b}(y_b, 1) & \mathbb{C}_{y_1,y_b}(y_b, 2) & ... & \mathbb{C}_{y_1,y_b}(y_b, k)$$ \\
		$$\mathbb{C}_{y_1,y_{b+1}}(y_{b+1}, 1) & \mathbb{C}_{y_1,y_{b+1}}(y_{b+1}, 2) & ... & \mathbb{C}_{y_1,y_{b+1}}(y_2,k)$$ \\
		\vdots \\
		$$\mathbb{C}_{y_1,y_{N-1}}(y_{N-1}, 1) & \mathbb{C}_{y_1,y_{N-1}}(y_{N-1}, 2) & ... & \mathbb{C}_{y_1,y_{N-1}}(y_{N-1},k)$$ \\
		$$\mathbb{C}_{y_1,y_N}(y_N, 1) & \mathbb{C}_{y_1,y_N}(y_N, 2) & ... & \mathbb{C}_{y_1,y_{N-1}}(y_N,k)$$ \\
	\end{Bmatrix}
	\]
	
	Note that rainfall for the year $y_i$ is  given by
	$$J_{y_i}= \sum_{l=1}^{k} \mathbb{C}_{y_1, y_i}(y_i,l) $$
	
\begin{figure}[H]
		\begin{center}
			\includegraphics[width=1.0\textwidth]{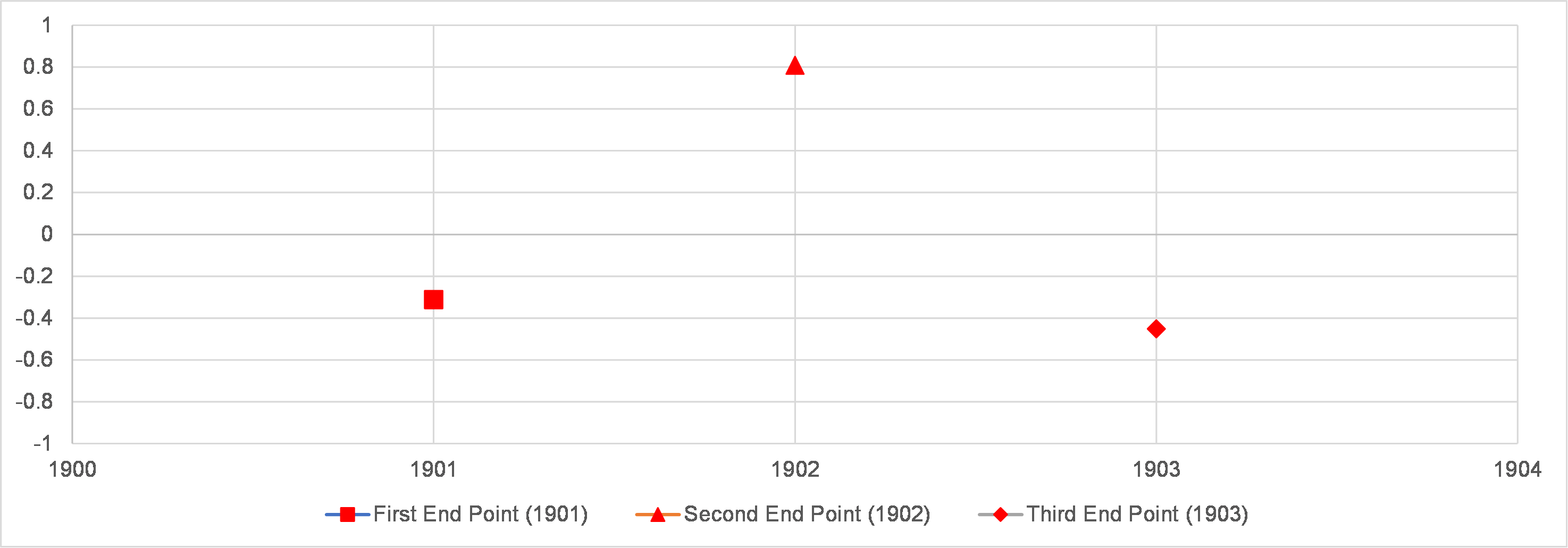}
			\caption{Variation of a component series denoted by WL8}  with Progressive Decomposition, Starting with 1901 and ending at 1903
			\label{fig:WL8-End-Points}
		\end{center}
	\end{figure}

	Elements of the above matrix are
	constant during the entire period
	of JJAS data, as they represent
	only the endpoints generated by the latest decompositions. Further,
	if a data point for the next year $y_{N+1}$ is added to JJAS, and
	it is decomposed into components, a new row with elements
	$\mathbb{C}_{y_1,y_{N+1}}(y_{N+1}, l)$ will be added but all the other
	elements of the above matrix will not change.  Each column of the
	matrix is like any conventional time series over the period chosen.
	Such time series is pictorially represented in Figure \ref{fig:WL8-End-Points},
	where the ending points of the MFs for WL8 are captured for three years.
	
	This formulation does not suffer from any of the flaws mentioned
	before, and the data set can be used to train, test and forecast in the
	the usual way, without having to worry about leakage of future data. So,
	this new method applies to other decomposition methods as well,
	{\it e.g.,} CEEMDAN, when applied to time series.
	It is worth describing an alternative view of the moving 
	front method of data collection. It is generally understood that
	the testing of a model can be considered objective if data in
	the training set used to develop the model are unbiased by the data in
	the testing set. As shown above, the coefficients $\mathbb{C}_{y_1,
		y_i}(y_i,l), b \leq i \leq n$ in the training data set are unaffected by
	the coefficients $\mathbb{C}_{y_1, y_k}(y_k,l), n+1 \leq k  \leq
	N$ in the testing set. Thus, one can expect moving front mode of data
	collection from progressive decomposition to make testing effective.
	
	\subsection*{Nature of Information in the Endpoint Matrix}
	It is to be noted that when JJAS data during the period $y_1$ to $y_N$
	is decomposed into component  series, the matrix $\mathbb{C}_{y_1, y_N}$ will
	have $(y_N - y_1) \times k$ elements. $\mathbb{E}_{y_b, y_N}$, the
	matrix generated by storing endpoints, has less number of
	elements, though the difference can be minimized by choosing the least
	possible value for $y_b$. More importantly, however, the nature of the information contained in the two
	matrices is different and richer in the latter. The element
	$\mathbb{C}_{y_1, y_N}(y_{m}, l)$ of the former matrix represents the
	value of the $l^{th}$ component  of the series corresponding to the year $y_m$ in
	its evolved state in the year $y_N$. The corresponding element in the
	$\mathbb{E}_{y_b, y_N}$ matrix is $\mathbb{C}_{y_1, y_m}(y_{m}, l)$. It
	on the other hand represents the value of the $l^{th}$ component of the series
	corresponding to the year $y_m$ as it was in its state evolved in the
	year $y_m$.  Thus, $\mathbb{E}_{y_b, y_N}$ is recording
	information of values in states evolved up to some year that lies in
	the period $y_b$ to $y_N$. The information contained
	in $\mathbb{E}_{y_b, y_N}$ is richer but the hope is that a DNN
	will be capable of learning the hidden patterns. The gain compensating
	for the additional complexity is the constancy of the elements
	to the addition of data points, which makes it very amenable to
	forecasting/predicting. It may also be noted that successive decompositions
	beginning from year $y_b$ are needed to generate the endpoint matrix
	and hence involves more effort. However, this extra effort is
	far less compared to that expended in developing entirely new models as practised in
	walk-forward validation methods, to be discussed later.
	
	\subsection*{The Choice of Deep Learning Network}
	A Deep Learning Neural Network (DNN) has more than one hidden layer,
	enabling them to process much more information, and hence it was chosen in this work. 
	
	Rainfall is a time series data, and the type of neural
	network selected has to recognize the time dependency. 
	Recurrent Neural Network (RNN) attempts to do this by taking into consideration
	the output of the previous time steps also, in addition to the
	customary input for the current time step. Though
	in theory, such pristine RNNs can learn arbitrarily long
	dependencies present in the input sequence, in practice, retaining
	long-term memory poses a problem with the same. Long Short-Term
	Memory networks have been developed as an alternative and are used in
	the present work. A brief description of LSTM is presented in Supplementary
	information B.
	
	\subsection*{Training, Testing and Forecasting}
	Consider the JJAS data from year $y_1$
	to year $y_N$. As mentioned earlier, some time is needed to begin
	decomposition into component series. Let the decomposition start from the year
	$y_b$. Thus, matrix $\mathbb{E}_{y_b, y_N} $ can be
	formed as described before.  Let the data series formed by the elements of columns of that
	matrix be
	divided into training set from $y_b$ to $y_{TR}$ and a testing set from
	$y_{TR+1}$ to $y_{TE}$. The rest of the data, i.e. $y_{TE+1}$ to $y_N$, are set aside for one-time forecasting and forecasting using Walk Foreward Validation (WFV) It is once again emphasized that elements of the submatrix
	$\mathbb{E}_{y_b, y_{TR}}$ are not influenced in any way by the elements
	of the submatrix   $\mathbb{E}_{y_{TR+1}, y_{TE}}$. 
	Therefore, the conventional method of splitting the data
	into training and testing, sets provide a valid test of the model. 
	Each of the $k$ data series will show patterns of
	variation. If these patterns could be learnt by an ML model, they can be used to make
	predictions of rainfall for each year in the testing period As $\mathbb{E}_{y_{TE+1}, y_{N}}$ are kept
	separate from both
	testing and training, and are independent of the other elements, they can be
	used for evaluating the effectiveness of a developed model in forecasting as well.
	
	\subsection*{The Calculation Scheme of LSTM}
	The standard multivariate LSTM is implemented, which is diagrammatically in Supplementary Material D.
	
	\subsection*{Computational Specifics of Predictions by EWT-MF-LSTM}
	Each column of the matrix $\mathbb{E}_{y_b, y_N}$ are the sub-series that are used as independent variables. The matrix $\mathbb{P}_{y_b, y_N} = $
	is obtained by combining with the $\mathbb{J}$ vector (dependent variable): 
	
	\[
	\mathbb{P}_{y_b, y_N} = 
	\begin{Bmatrix}
		$$\mathbb{C}_{y_1,y_b}(y_b, 1) & \mathbb{C}_{y_1,y_b}(y_b, 2) & ... & \mathbb{C}_{y_1,y_b}(y_b, k), \ J_{y_1}$$ \\
		$$\mathbb{C}_{y_1,y_{b+1}}(y_{b+1}, 1) & \mathbb{C}_{y_1,y_{b+1}}(y_{b+1}, 2) & ... & \mathbb{C}_{y_1,y_{b+1}}(y_2,k, \ J_{y_2}$$ \\
		\vdots \\
		$$\mathbb{C}_{y_1,y_{N-1}}(y_{N-1}, 1) & \mathbb{C}_{y_1,y_{{N-1}}}(y_{N-1}, 2) & ... & \mathbb{C}_{y_1,y_{N-1}}(y_{N-1},k), \ J_{y_{(N-1)}}$$ \\
		$$\mathbb{C}_{y_1,y_N}(y_N, 1) & \mathbb{C}_{y_1,y_N}(y_N, 2) & ... & \mathbb{C}_{y_1,y_{N-1}}(y_N,k), \ J_{y_N}$$ \\
	\end{Bmatrix}
	\]
	
	These parallel time series, i.e. the columns in the matrix $\mathbb{P}_{y_b, y_N}$ were re-framed as a
	supervised learning problem by transforming the sequence in the
	form of input and output pairs using lag value. As mentioned
	earlier, some finite period is needed to create a component of the series, and end-point
	collection can start. Thus, 1901 was chosen as the starting point.
	Out of available data (1871-2022), eight component series and one residue were
	extracted by progressive decomposition using EWT-MF, ultimately forming nine ($k$ = 9) parallel series of end data points,
	the points for the period 1901-1980 were 
	devoted to training and the next 1981 to 1999 as testing. The data from 2000 to 2022 were reserved for forecasting.  
	
	As mentioned before, a lag
	period ($\mathit L$) or window size was used in re-framing the data $\mathbb{P}$ into a supervised learning problem. Different window sizes, number of hidden layers, number of nodes in each layer, learning
	rate, tolerable number of epochs to allow a monotonic increase of error before the training is stopped, batch size and so on 
	are varied interactively using experience. The mentioned parameters are called {\it hyperparameters} and their values are not known {\it a-priori} and must be found by trial and error. Grid search for the above parameters was not conducted, as it is computationally very expensive.
	The model is trained by using EWT-MF-LSTM data during the period 1901-1980. 
	
	For each row of the supervised data of independent variable (in a pre-defined tensor format) input to LSTM, it makes predictions for rainfall for each row. When the last $\mathit L$ rows of supervised data of independent variable (again in a particular tensor format), are input to LSTM, it makes one unknown future forecast for rainfall. 
	
	The testing period was defined as 1981-1999. The trained model was evaluated against the test set. If the error between actual and predicted values is tolerable, the model is frozen. Otherwise, the model is re-built with modified hyperparameters and the valuation is done against the testing set.
	
	Next, the satisfactory model obtained from the above step is trained for the entire data from 1901-1999, keeping the model hyperparameters to be the same, to arrive at the model that is ready for forecasting from 2000-2022. It is customarily referred to as the final model.
	
	\subsubsection*{Stochastic Nature}
	ANN solutions are non-deterministic because they use random
	numbers for (a) the initialization of weights attached to each node of
	the network (b) the optimization of weights to solve by
	Gradient Descent method. This implies that
	a different pathway will be followed when the same algorithm with
	the same data is repeatedly run, resulting in a slightly different 
	solution. The randomness can be stopped by specifying a random
	seed. However, the recommended
	practice in Data Science is to run the same program with the same
	data an appreciable number of times and then calculate the
	result as the average of all the runs. The number of repeated runs
	has been chosen as 20 since beyond that there is no appreciable
	difference.
	
	\subsubsection*{Programming Language and Packages Used}
	As mentioned earlier LSTM was the choice of deep learning to deal with the high variability of rainfall data. To this end, a description of the programming language and the packages used are described here.
	{\it{Python}} 3.9 was used to develop the computer program utilizing
	the available packages: {\it{numpy}}, {\it{pandas}}, {\it{numpy}},
	{\it{sklearn}}, {\it{pickle}}, {\it{statsmodels}}, {\it{matplotlib}},
	{\it{seaborn}}, {\it{statistics}}. 
	
	Python package of EWT (ewtpy) is available at
	\url{https://pypi.org/project/ewtpy/}. {\it{Keras}},
	an open-source deep-learning Application Programming Interface (API)
	(\url{https://pypi.org/project/keras/})
	written in {\it{Python}} was used to train and predict LSTM
	networks. {\it{Keras}} acts as an interface for the {\it{TensorFlow}}
	(\url{https://www.tensorflow.org/install/pip})
	library, developed by Google Brain. The versions of {\it{Keras}} and
	{\it{Tensorflow}} were 2.4.3 and 2.4.0 respectively.
	
	\section*{Results \& Discussion}
	The section is divided into three parts. We begin by presenting
	results obtained from the application of the LSTM model developed in the
	present work. It is followed by a discussion of other models
	reported in the literature. Finally, predictions by statistical
	methods are presented. 
	
	\subsection*{Predictions of EWT-MF-LSTM model}
	The Moving Front (MF) data started in 1901. Table \ref{tab:TRAINING-TESTING-FORECASTING} shows the duration of the training, testing and forecasting (WFV without and with retraining).
	
	The year ranges are shown in Table \ref{tab:TRAINING-TESTING-FORECASTING}. 
	\begin{table}[H]
		\centering
		\caption{Training, Testing and Forecasting Periods}
		\begin{tabular}{|l|l|l|}
			\rowcolor[rgb]{ .753,  0,  0} \textcolor[rgb]{ 1,  1,  1}{\textbf{Training}} & \textcolor[rgb]{ 1,  1,  1}{\textbf{Testining}} & \textcolor[rgb]{ 1,  1,  1}{\textbf{Forecasting}} \\
			\hline
			1901-1980 & 1981-1999 & 2000-2022 \\
			\hline
		\end{tabular}%
		\label{tab:TRAINING-TESTING-FORECASTING}%
	\end{table}%
	
	The corresponding $PP$ values are listed in Table \ref{tab:PP}.
	
	\begin{table}[H]
		\centering
		\caption{$PP$ Values}
		\begin{tabular}{|r|r|r|r|}
			\rowcolor[rgb]{ 0,  0,  .4} \multicolumn{1}{|l|}{\textcolor[rgb]{ 1,  1,  1}{\textbf{Training}}} & \multicolumn{1}{l|}{\textcolor[rgb]{ 1,  1,  1}{\textbf{Testining}}} & \multicolumn{2}{c|}{\textcolor[rgb]{ 1,  1,  1}{\textbf{Forecasting}}} \\
			\cmidrule{3-4}    \rowcolor[rgb]{ 0,  0,  .4} \textcolor[rgb]{ 1,  1,  1}{} & \textcolor[rgb]{ 1,  1,  1}{} & \multicolumn{1}{c|}{\textcolor[rgb]{ 1,  1,  1}{\textbf{WFV (No Retraining)}}} & \multicolumn{1}{c|}{\textcolor[rgb]{ 1,  1,  1}{\textbf{WFV (Retraining)}}} \\
			\hline
			0.99  & 0.86  & 0.90  & 0.95 \\
			\hline
		\end{tabular}%
		\label{tab:PP}%
	\end{table}%
	
\begin{figure}[H]
		\begin{center}
			\includegraphics[width=1.0\textwidth]{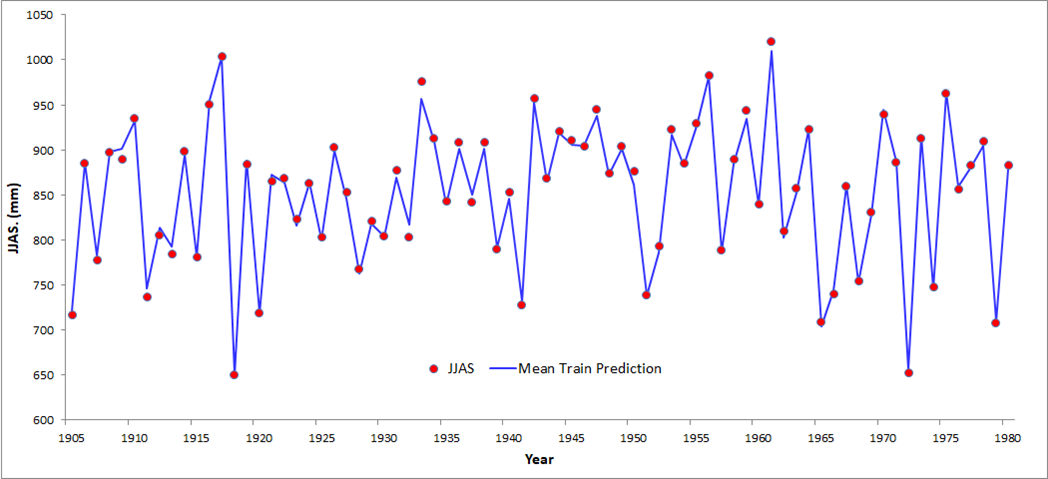}
			\caption{Observations and model predictions during training period}
			\label{fig:Training}
		\end{center}
	\end{figure}

\begin{figure}[H]
		\begin{center}
			\includegraphics[width=1.0\textwidth]{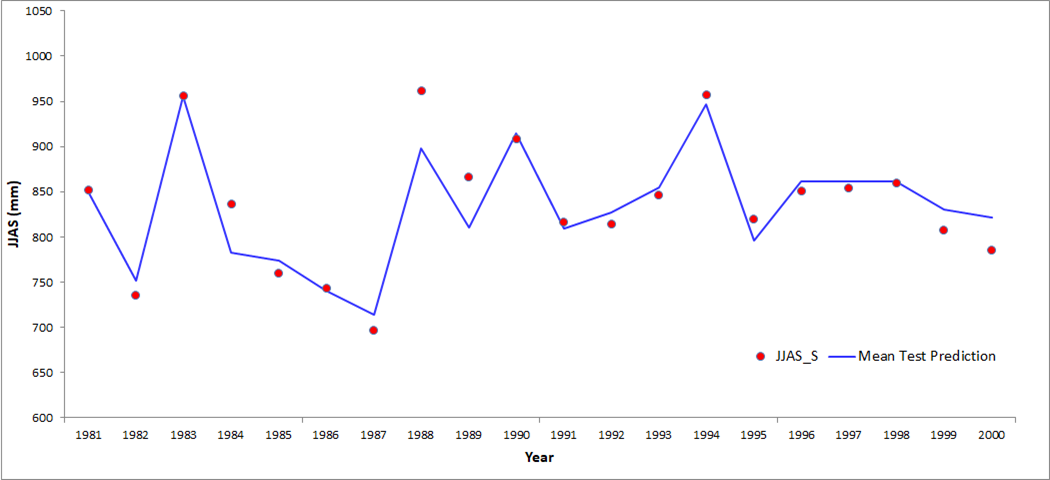}
			\caption{Observations and model predictions during the period assigned for testing.}
			\label{fig:Testing}
		\end{center}
	\end{figure}

\begin{figure}[H]
		\begin{center}
			\includegraphics[width=1.0\textwidth]{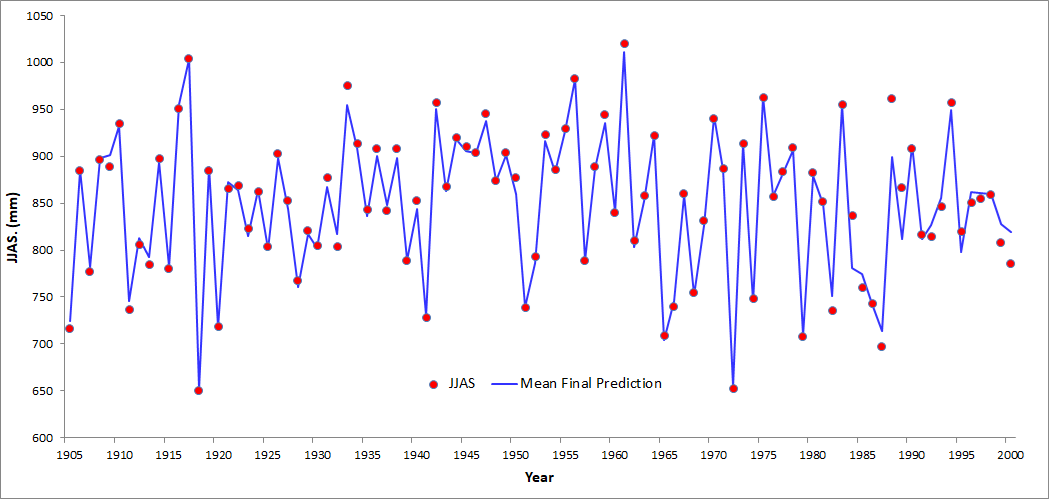}
			\caption{Observations and predictions of the final model for the training and testing period.}
			\label{fig:Final}
		\end{center}
	\end{figure}

	The observed versus predicted values are plotted in Figure \ref{fig:Training}, \ref{fig:Testing} and \ref{fig:Final}, for training, testing and the final models. 
	
	The red points are the observed values, and the Blue lines represent predictions. The reported value is obtained by repeating the same experiment 20 times
	as discussed before.

	\subsection*{Forecasting and Comparison with the Other Models}
	
\begin{figure}[H]
		\begin{center}
			\includegraphics[width=1.0\textwidth]{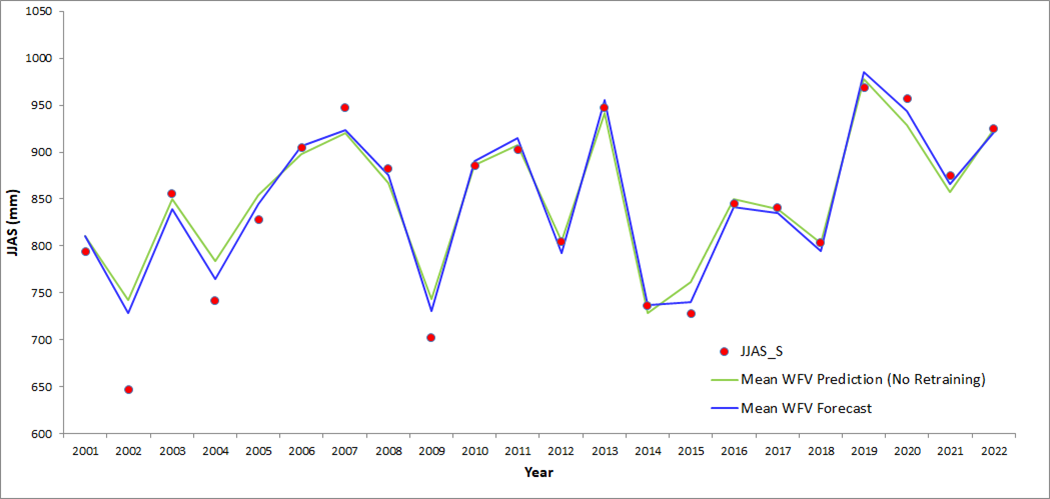}
			\caption{WFV Forecasts without and with retraining}
			\label{fig:WFV}
		\end{center}
	\end{figure}

	As discussed earlier, when multiple rows of the supervised data, in a compatible tensor format is input to the trained (or final) model, the same number of predictions/forecasts for $JJAS$ are obtained. Thus, if the data from 2000-2022 are fed to the final model (which has been trained using data from 1901-1999), 23 years of forecasts are obtained, at a time. In Figure \ref{fig:WFV}, the Green line represents these forecasts. The $PP$ of these forecasts were 0.90.
	It is to be noted that this is an in-sample forecast and serves the purpose of determining the efficacy of the learning by DNN, without the influence of training and testing data set. Since $JJAS$ is already known for the period 2000-2022, and hence the EWT-MF points, it cannot be called a true forecast.
	
	Another way suggested uses methods that are
	referred to in general as Walk Forward Validation (WFV)
	techniques. In the most widely practised
	method, the model is
	trained with all the data up to the latest available to make a forecast for the
	next year.  When an observation for this forecast
	becomes available next year, the model is updated to include that
	value, and a forecast is made for the following year. Hence the
	name WFV. This process yields a forecast, one year at
	a time. This method is followed in this work for the same forecasting period (2000-2022). The WFV forecasts are presented in Figure \ref{fig:WFV} in colour Blue. The attained $PP$ = 0.95 is higher than when WFV was not used and can be considered very good.
	
	When WFV, with retraining at each step, is applied using EWT-MF forecast data from 2000-2022, the LSTM has a better chance to learn the variations of the EWT-MF components, since it is progressively retrained with the addition of the forecast data one at a time. This variation can be learnt during the training phase, but not necessarily completely, as the model becomes more and more skilful with the added data. For retraining progressively by adding forecast data one by one, {\it all other hyperparameters are kept the same, i.e. no new model was generated.}
	It is pertinent to add that, instead of feeding all 23 years of supervised data at once, one set of supervised rows can be input at a time, just like WFV, but {\it no retraining is done}. This also results in the same output values as before (23 years at a time) with a $PP$ of 0.90.
	
	Earlier work on different forecasting strategies that used decomposition techniques is discussed here. It is difficult to compare the present work with others
	since decomposition strategies and neural network models are
	different. The comparison is presented to simply give an idea of the
	relative performances.
	
	Some simpler routes have been tried to avoid the leakage of future
	information. In one such, all the available data are  decomposed once only, and the constituent series are trained
	and used to make predictions, {\it e.g.,} JJAS data from 1901-2022 
	followed by a forecast for 2023.
	However, the reliability of such a model is unknown in the absence of
	any evaluation procedure. Further, a model developed by minimising
	only the training error often results in an over-fitted model. 
	
	\citet{Iyengar2003} divided JJAS data into training and forecast
	sets. Forecasts were made by a combination of regression and the walk-forward method.   Details of training have not been provided, but they
	report a $PP$ of 0.83 for training. This may be compared 
	with a $PP$ of 0.90 reached during testing in this work. It is also worth noting that
	they attained a $PP$ of 0.82 for the forecast set, which is nearly the same
	as that for training. This is a much poorer performance
	compared to that obtained for forecast in the present work.
	
	\citet{Johny2020} developed another variation of WFV. After dividing the data into training
	and testing, the training data is divided into a quasi-training and
	quasi-testing set. A model is developed ignoring the fact that both
	the quasi-train and quasi-test set will contain future
	information. They then proceed to use the saved trained model to make
	a forecast for one year. Redecomposition is made after including the actual value of the forecasted year, followed by retraining and another forecasting. As the saved model is used, i.e. no change in the model is done, though test data with data leak are
	being used, it is the closest to compare with the present work.  
	They do not
	report $PP$ for the forecast period. However, the Mean Absolute Error
	reported can be used in place of RMSE to estimate $PP$. It is
	about 0.8, lower than what was obtained in this
	work. \citet{Johny2020} also used another strategy, referred to as AEEMD,  where an entirely new
	model is developed every time a data point is added from the test set, to make
	forecasts. A $PP$ of these forecasts is estimated from
	their parity plots to be 0.75, a value
	near that of $PP$ for forecasts. 
	
	Predictive/forecasting capability of the present
	EWT-MF-LSTM method is compared with that of others in Table \ref{tab:comp}. Here the performance numbers quoted are as reported
	by the respective authors and, if it was not reported by the authors
	it was computed from the values provided by them where possible.  As
	can be seen, the present method does achieve better accuracy than other methods.
	
	\begin{sidewaystable}[h!]
		\centering
		\caption{Broad comparison of the performance of predictive capabilities of models}
		{\scriptsize
			\begin{tabular}{|l|c|c|c|l|c|c|c|c|c|c|}
				\hline
				\multicolumn{1}{|p{6.055em}|}{Reference}   & \multicolumn{1}{p{3.39em}|}{Rainfall Data} & \multicolumn{1}{p{3.39em}|}{Training}  & \multicolumn{1}{p{3.39em}|}{Testing} & \multicolumn{1}{p{5.39em}|}{Method} & \multicolumn{1}{p{2em}|}{Future Prediction} & \multicolumn{1}{p{1.89em}|}{RMSE (mm)} & \multicolumn{1}{p{2.835em}|}{NRMSE} & \multicolumn{1}{p{2.055em}|}{MAPE} & \multicolumn{1}{p{1.39em}|}{R} & \multicolumn{1}{p{1.39em}|}{$PP$} \\
				\hline
				{\bf Hybrid models: present and others}& \multicolumn{10}{l|}{}\\
				\hline
				Present Work  & 1871-2022 & 1901-1999 &   2000-2022 & EWT-MF-LSTM &  & 4.49 & 0.005 & 0.0039  & 0.997  & 0.95 \\
				\hline
				\citet{Johny2020}   & 1871-2018 & 1871-1970        & 1971-2018 & AEEMD-ANN &     &       & 0.14  &       & 0.91  &  0.75\\
				\hline
				\citet{Iyengar2003}    & 1871-1994 & 1872-1990        & 1991-2003 & 	EMD-NN-Regression & 1 year &       &       &       &  & 0.82 \\
				\hline
				\multicolumn{11}{l}{\tiny} \\
				\multicolumn{11}{l}{\tiny * Acronyms:} \\  
				\multicolumn{11}{l}{\tiny \ EWT: Empirical Wavelet Transform} \\ 
				\multicolumn{11}{l}{\tiny \ MF: Moving Front} \\ 
				\multicolumn{11}{l}{\tiny \ LSTM: Long Short-Term Memory} \\   
				\multicolumn{11}{l}{\tiny \ AEEMD: Adaptive Ensemble Empirical Mode Decomposition} \\
				\multicolumn{11}{l}{\tiny \ ANN: Artificial Neural Network} \\   
				\multicolumn{11}{l}{\tiny \ EMD: Empirical Mode Decomposition} \\
			\end{tabular}
			\label{tab:comp}
		}
	\end{sidewaystable}
	
	\section*{Conclusions}
	Complex time series have very high variability, and due to their highly
	non-linear complex nature, not even DNN alone may be insufficient to learn
	and forecast it with high accuracy. In such instances, many
	researchers in different domains employed the decomposition of the
	parent time series into simpler constituent series using signal
	processing tools like EMD, EEMD, CEEMDAN, EWT and VMD. The idea is to
	learn the sub-series individually and then sum up the individual
	predictions to reproduce the original time series or to formulate a multivariate problem.  Decomposition of
	the whole dataset only once and dividing it into training and testing sets
	is widely practised in literature. However, a property of the one-time
	decomposition technique is that each of the sub-series changes its
	nature depending upon the starting and ending period. This leads to
	leakage of future information, making the whole testing process ineffective,
	and has been referred to as a 'data leak'. A novel MF method has been
	proposed in this work to prevent the data leak so that the decomposed
	constituent time series can be treated as normal time series and reap
	the benefits of decomposition as well. The technique employs progressive
	decomposition and collection of endpoints at each step.  ISMR data was selected to demonstrate the efficacy of the MF method. EWT was found
	to be superior to CEEMDAN and was used to decompose ISMR data
	employing the MF method. Instead of using the conventional univariate
	technique of learning decomposed time series individually, a
	multivariate formulation is employed in the present work. Here, the
	constituent series form a parallel independent series with the target
	time series as the dependent one.  State-of-the-art LSTM network architecture with superior sequence
	predictive capabilities was used in this work. The entire data was
	divided into training, testing and forecasting periods. The performance of
	the resulting MF-EWT-LSTM model was excellent in training and testing.
	Forecasts were obtained by a finalized model all at once and also
	using a walk-forward method widely used in the literature. The results
	obtained were much superior to those reported in the literature.

	\clearpage
	
	\section*{Supplementary Material}
	\subsection*{A. Statistical Properties}
	\setcounter{table}{0}
	\setcounter{figure}{0}
	The probability density function revealed a skewed
	distribution. Almost no (or very weak) autocorrelation was
	found, and this explains why the linear time-series modelling
	failed. "Stationarity" has to be ascertained, and any trend and/or seasonal
	effects are to be removed before beginning neural network modelling of
	time-series data. The {\it{seasonal\_decompose}}
	the function of the Python package {\it{statsmodels}} has been used to
	determine "seasonality" (a data science term), it was found
	seasonal component was absent. This was to be
	expected as the data was the seasonal average.
	
	The Augmented Dickey-Fuller (ADF) test by \citet{SAID1984} is a common
	statistical test to determine how strongly a time series is defined by
	a trend.  The result of this test is interpreted using the
	$p$-value.   Using a Python function
	{\it{adfuller}} from the {\it{statsmodels}} package, the $p$-value was
	found to be zero for the JJAS, and it is inferred that the time series
	is stationary.  Hence, the data, as retrieved, were used as the input
	to the forecasting models.
	
	\subsection*{B. Long Short-Term Memory (LSTM)}
	RNNs using Long Short-Term Memory (LSTM) units can
	decide whether to keep the long-term memory or not. LSTMs use a
	series of gates (Forget Gate, Input Gate and Output Gate) and a new-memory block which
	controls how the sequential information enters and leaves the network
	while retaining the relevant {\it memory}. This enables LSTM to allow
	the previously mentioned gradients to flow unchanged, thereby
	partially solving the vanishing gradient problem.  LSTM was proposed
	by \citet{HochSchm97}. Google, Apple, Amazon, Facebook and Microsoft
	are among the giant companies that use LSTM, and it, therefore, is a
	time-tested neural network with a significant ability to predict
	sequential data. It is chosen in this work also.  
	The gates in the LSTM mentioned and the New Memory block is all neural networks in themselves and all of them receive the same two inputs: the previous hidden state
	$h_{t-1}$, and input for the current step $x_t$. The outputs of the
	Forget, Input and Output are $f$, $i$ and $o$, suffixed by the time step
	$t$. 
	At the very end, the hidden state $h_t$ is converted to the final
	result by a linear layer to give the actual predicted value
	$\hat{y}_t$:
	
	$$ \hat{y}_t = W  h_t $$
	
	The equations below, for the forward pass, are presented below. The influence of
	{\it{bias}} has not been taken into account. 
	
	Equations for the forward pass of LSTM are:
	\begin{align*}
		f_t &= \sigma(W_{f} x_t + U_{f} h_{t-1}) \\
		i_t &= \sigma(W_{i} x_t + U_{i} h_{t-1}) \\
		o_t &= \sigma(W_{o} x_t + U_{o} h_{t-1}) \\
		\tilde{c}_t &= \sigma(W_{c} x_t + U_{c} h_{t-1}) \\
		c_t &= f_t \circ c_{t-1} + i_t \circ \tilde{c}_t \\
		h_t &= o_t \circ \sigma_h(c_t)
	\end{align*}
	where the initial values are ${\displaystyle c_{0}=0}$ and
	${\displaystyle h_{0}=0}$ and the operator ${\displaystyle \circ }$
	denotes the element-wise product. The subscript ${\displaystyle t}$
	indexes the time step. Further, \\
	
	\noindent ${\displaystyle x_{t}}$ : Input vector to the LSTM unit \\
	${\displaystyle f_{t}}$ : Output vector of the Forget Gate \\
	${\displaystyle i_{t}}$ : Output vector of the Input Gate \\
	${\displaystyle o_{t}}$ : Output vector of the Output Gate \\
	${\displaystyle h_{t}}$ : Hidden state vector is also known as the output vector of the LSTM unit \\
	${\displaystyle {\tilde {c}}_{t}}$ : Output vector of the New memory cell\\
	${\displaystyle c_{t}}$ : Cell state vector \\
	\\
	\noindent ${\displaystyle W}$, ${\displaystyle U}$ : Weight matrices which are adjusted during training \\
	
	\noindent The Activation functions are:\\
	
	${\displaystyle \sigma}$ : Sigmoid function \\
	${\displaystyle tanh}$ : Hyperbolic tangent function \\
	
	\subsection*{C. Comparison of Data Decomposition Methods}
\begin{figure}[H]
		\begin{center}
			\includegraphics[width=0.45\textwidth]{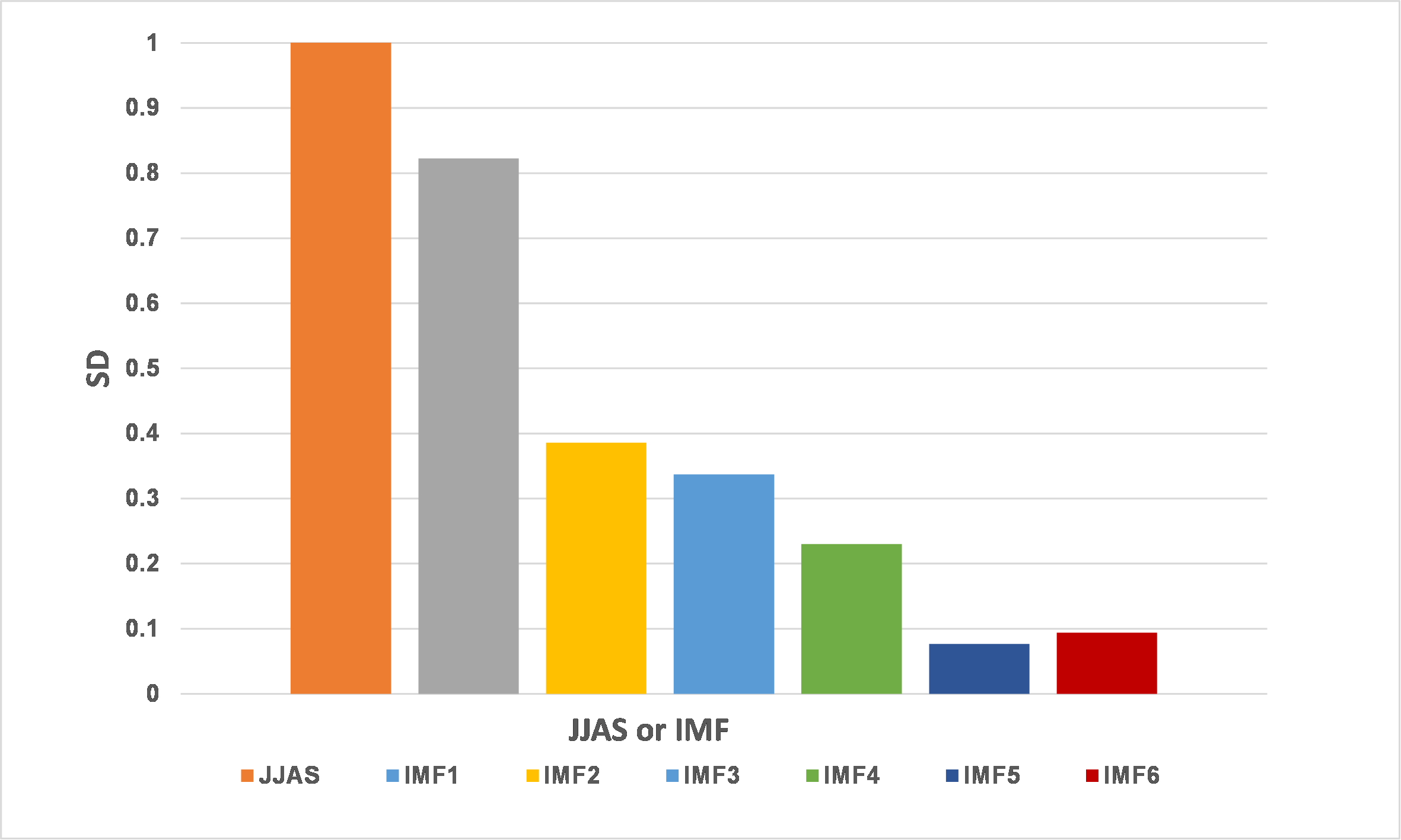} \includegraphics[width=0.45\textwidth]{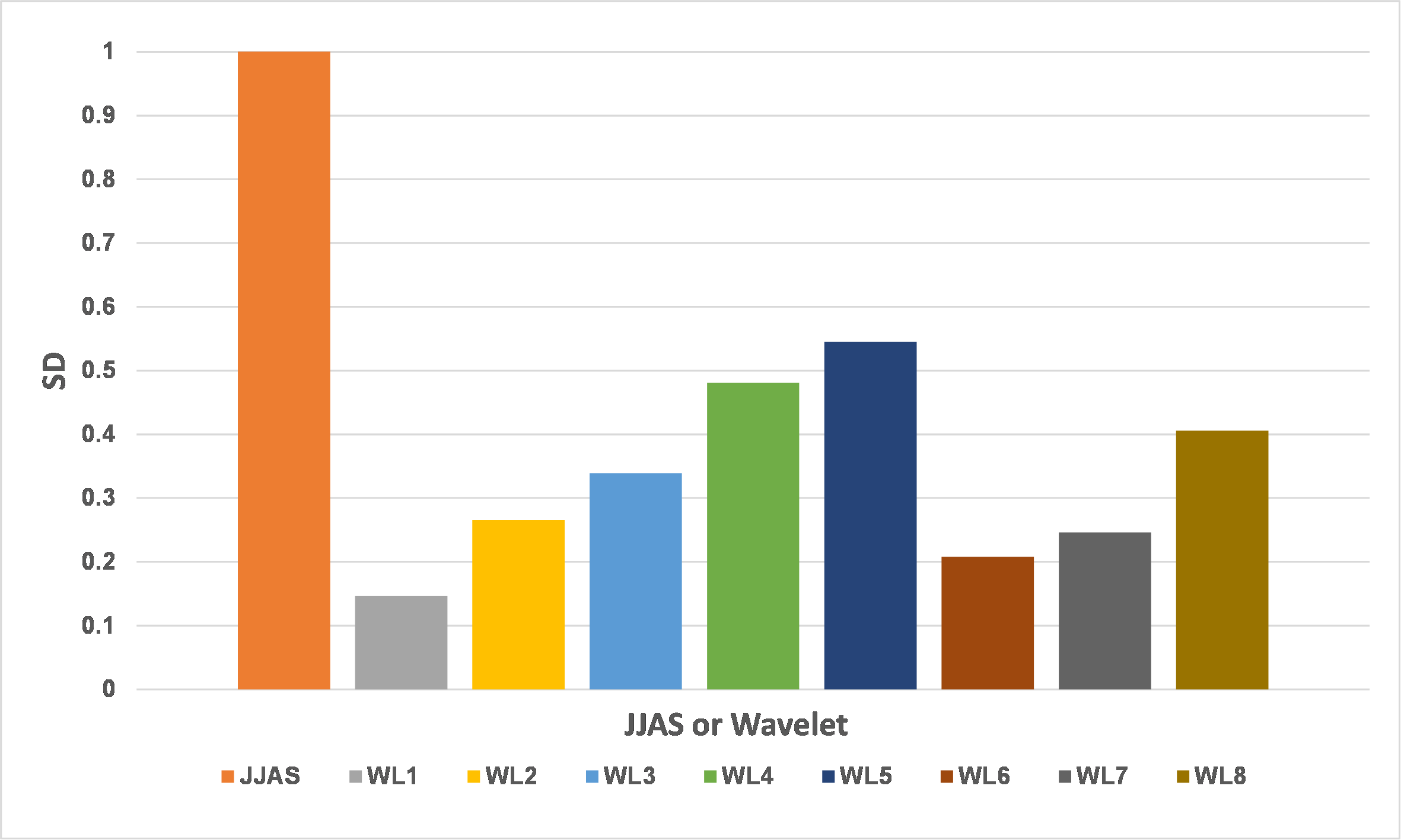}
			\caption{Left panel shows a comparison of Standard Deviations (SD)
				of JJAS with that of the IMFs extracted using CEEMDAN. Right panel
				shows a comparison of Standard Deviations (SD) of JJAS with that of Wavelets extracted using EWT.
				\label{fig:sd-ceemdan-imfs-ewt-waves}}
		\end{center}
	\end{figure}

	As discussed before, CEEMDAN is the most effective technique in the EMD family, and this was chosen to decompose JJAS data. 
	However, with the LSTM network and CEEMDAN decomposition described above, the predictive capability was not satisfactory.
	This prompted an investigation to find
	out possible reasons for the lack of improvement. The first IMF (IMF1)
	computed by CEEMDAN appeared to account for most of the
	non-linearity of JJAS as indicated by the
	graphs of the IMFs.  The comparison is shown in
	the left panel of Figure
	\ref{fig:sd-ceemdan-imfs-ewt-waves}.
	
	It is evident from the figure that IMF1 contained the majority
	of the high  amplitude components, making it difficult to learn by
	ANN. A better distribution is not possible since the EMD/EEMD/CEEMDAN
	algorithms fixed the number of IMFs automatically determined.  Lack of
	improvement in $\mathit{PP}_{Test}$ can be attributed to the inability of the CEEMDAN
	algorithm to distribute the variances in the original signal into a
	large number of less complex components. 
	Secondly, it was observed that the LSTM network
	could not predict  IMF1 with the desired accuracy. This indicated the complexity of the patterns in IMF1 was so
	high that it was not amenable even to the sophisticated deep learning
	LSTM networks. 
	
	As mentioned earlier, a
	higher number of modes can be specified with EWT. Hence, EWT was
	tried as an alternative to decompose JJAS.  The right panel of Figure
	\ref{fig:sd-ceemdan-imfs-ewt-waves} shows the SD for different wavelets obtained
	versus the SD of the rainfall data, after extracting 8 modes by
	EWT. It is evident from Figure \ref{fig:sd-ceemdan-imfs-ewt-waves}, the fluctuations in JJAS are more evenly
	distributed among the decomposed time series for EWT as shown in Figure \ref{fig:sd-ceemdan-imfs-ewt-waves}.
	
	\subsection*{D. Calculation Scheme}
\begin{figure}[H]
		\begin{center}
			\includegraphics[width=0.9\textwidth]{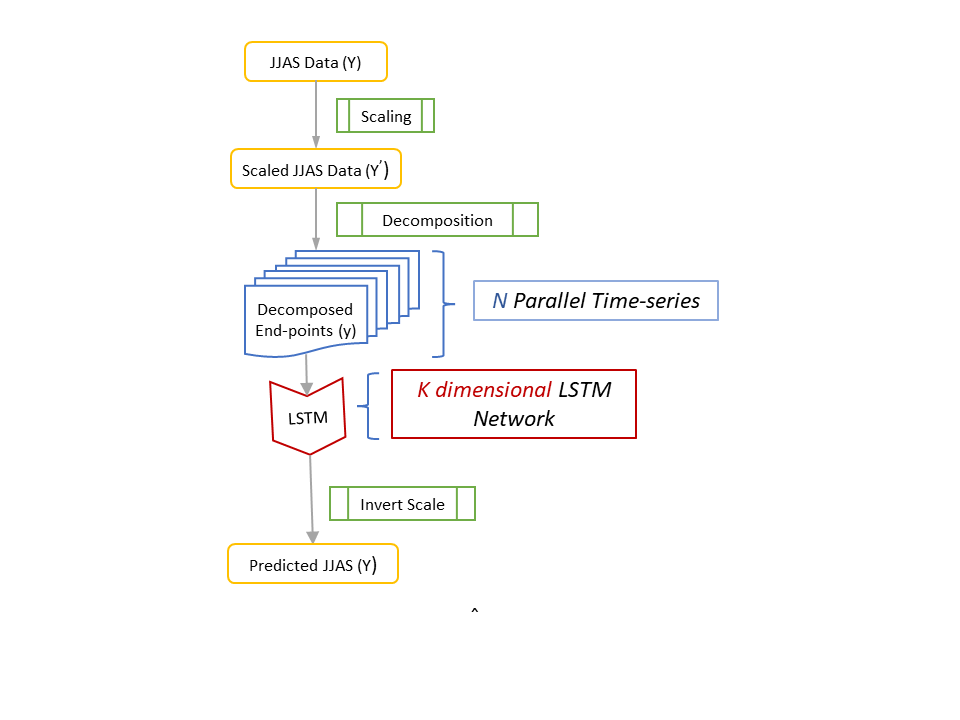}
			\caption{The calculation flow diagram
				\label{fig:flow}}
		\end{center}
	\end{figure}

	Figure \ref{fig:flow} shows the flow diagram. The JJAS
	data, first scaled using the mean and the SD, was decomposed by the
	decomposition algorithm (EWT) into the $k$ constituent simpler
	time series, following the MF method described earlier. Each such time series was input to $k$ dimensional multi-variate LSTM network, for
	training and making predictions. The prediction is inversely scaled to get
	the predicted JJAS.
	
	\clearpage
	\section*{Statements and Declarations}
	\begin{itemize}
	   \item Acknowledgements: The author sincerely thanks Profs. J. Srinivasan of Divecha Center for Climate Change, Indian
		Institute of Science, Bangalore, and K.S. Gandhi, Department of Chemical Engineering, Indian
		Institute of Science, Bangalore, for insightful discussions and
		critical comments on the manuscript. Also, I thank Prof. K.V.S. Hari, Department of ECE,
		Indian Institute of Science, Bangalore, for his expert comments on the method proposed.
		\item Funding: The author declares that no funds, grants, or other support were received during the preparation of this manuscript.
		\item Competing interests: The authors have no competing interests to declare that are relevant to the content of this article.
		\item Compliance with Ethical Standards: The research meets all ethical guidelines, including adherence to the legal requirements of the study country.
		\item Consent to participate: Not applicable.
		\item Consent for publication: The author agrees with the content and gave explicit consent to submit and that he obtained consent from the responsible authorities at the Indian Institute of Science, where the work has been carried out before the work is submitted.
		\item Availability of data and materials: The rainfall data from the year 1871 to 2016 and 2017 to 2022 
		were obtained from the
		Indian Institute of Tropical Meteorology (IITM) and Indian Meteorological Department's (IMD) websites respectively.\footnote{Data were provided by Prof J Srinivasan, Divecha Center for Climate Change, Indian Institute of Science, Bangalore}
		\item Research involving Human Participants and/or Animals: Not applicable.
	\end{itemize}
	
	\clearpage
	
	\bibliographystyle{unsrtnat.bst}
	\bibliography{paper_rainfall.bib}
	
\end{document}